\documentclass[runningheads]{llncs}
\usepackage{graphicx}
\usepackage{tabularx}
\usepackage{booktabs} 
\usepackage{tablefootnote}
\usepackage{subcaption}
\usepackage{hyperref}
\usepackage{orcidlink}

\begin{document}
%


\title{Combining Knowledge Graphs and NLP to Analyze Instant Messaging Data in Criminal Investigations\thanks{This preprint has not undergone peer review or any post-submission improvements or corrections. The Version of Record of this contribution is published at WISE 2024, and is available online at \url{https://doi.org/10.1007/978-981-96-0567-5_30}.}}

%
\author{Riccardo Pozzi\inst{1}\orcidlink{0000-0002-4954-3837} \and
Valentina Barbera\inst{1}\orcidlink{0009-0000-5177-0919} \and
Renzo Alva Principe\inst{1}\orcidlink{0000-0002-8540-6389} \and
Davide Giardini\inst{1}\orcidlink{0009-0007-8909-2309} \and
Riccardo Rubini\inst{1}\orcidlink{0009-0001-0955-6721} \and
Matteo Palmonari\inst{1}\orcidlink{0000-0002-1801-5118}}
\authorrunning{R. Pozzi et al.}
%
\institute{University of Milano-Bicocca, Milan, Italy}
%
\maketitle              
\begin{abstract}
Criminal investigations often involve the analysis of messages exchanged through instant messaging apps such as WhatsApp, which can be an extremely effort-consuming task. Our approach integrates knowledge graphs and NLP models to support this analysis by semantically enriching data collected from suspects' mobile phones, and help prosecutors and investigators search into the data and get valuable insights. Our semantic enrichment process involves extracting message data and modeling it using a knowledge graph, generating transcriptions of voice messages, and annotating the data using an end-to-end entity extraction approach. We adopt two different solutions to help users get insights into the data, one based on querying and visualizing the graph, and one based on semantic search. The proposed approach ensures that users can verify the information by accessing the original data. While we report about early results and prototypes developed in the context of an ongoing project, our proposal has undergone practical applications with real investigation data. As a consequence, we had the chance to interact closely with prosecutors, collecting positive feedback but also identifying interesting opportunities as well as promising research directions to share with the research community.






\end{abstract}
\section{Introduction}
\label{sec:introduction}



In a criminal investigation, \textit{investigators} like prosecutors and law enforcers\footnote{In this paper we use the broader term \textit{investigators} to refer to prosecutors and personnel of various kinds of law enforcement agencies, usually coordinated by the prosecutors. Our goal is to abstract away from details about cooperation schemes in the investigations which also depend on country-specific legislations.} need to acquire and analyze evidence collected from a large number of heterogeneous sources, including structured, unstructured, and multimedia data~\cite{batini2021semantic}.

Considering the crucial role of entities in this domain, it is not surprising that semantic data integration and knowledge graphs~\cite{knoblock2015exploiting} (KGs) have been proposed as valuable assets to support investigative activities~\cite{szekely2015building,batini2021semantic}. Semantics is applied to interlink and organize data collected from heterogeneous sources, adopting an entity-centric data management approach. Even a lightweight use of KGs can provide several benefits to help investigators search over large amounts of data and get insights from it~\cite{szekely2015building}. Several companies operating with semantic models also target investigative applications, e.g., Siren\footnote{\url{https://siren.io/}}. Generative AI and Language Models (LMs) are expected to have a huge impact on data processing in this 
domain. However, we believe that these novel technologies, rather than wiping out the key benefits of semantics in a domain where data integration and organization play a crucial role, can unlock novel opportunities.

In this work, we focus on the specific problem of supporting the analysis of information extracted from Instant Messaging Apps (IMA), such as WhatsApp. The suspects' smartphones are often seized and examined during a criminal investigation and contain extremely valuable evidence. However, according to the investigators we collaborated with, a complete analysis of the data exchanged via IMAs on even a single smartphone is so effort-consuming that it is almost out of reach in most investigations. In two investigations considered in our work, more than one thousand chats were extracted from an IMA. As of today, the exploration of textual messages is supported by existing software products with reportedly limited capabilities (scrolling, syntactic search); the analysis of data exchanged through IMAs instead, especially voice messages, is even more demanding and not supported by any tool they have at their disposal. In the best cases, the time required to complete the analysis of IMA is severely impacted, in the worst cases, the effort is beyond the resources available. 




The work described in this paper is part of a larger national Italian project where two research teams coordinated by an inter-university institution (of which we do not report the name for double blindness) 
are experimenting with semantic data integration approaches to support criminal investigations~\cite{batini2021semantic}. As a research team, our mission in the project is to develop Proof-of-Concepts (PoCs) to evaluate the feasibility and usefulness of AI-based solutions in this domain and define key functionalities to be developed by an industrial subject into final software products. The more specific objectives of our activities in the domain of IMA analysis are not only to exploit the latest generation of deep learning-based speech-to-text technology to obtain transcripts of audio content in IMAs and evaluate their fitness for use in this domain, but also to integrate this into a semantic data integration paradigm, where chats (and transcripts) are further processed to extract entities, and integrated data made accessible with a semantic search application.
Our work is formalized under an official consultant contract, which allows us to develop PoCs, to create reports based on actual investigation data, and to collect feedback from investigators.
Access to data is covered by precise legislation that regulates the relationship between investigators and consultants.
While the project is ongoing, we believe that the activities carried out so far and the preliminary results are interesting to be shared with the research community. 

\begin{figure}
\includegraphics[width=\textwidth]{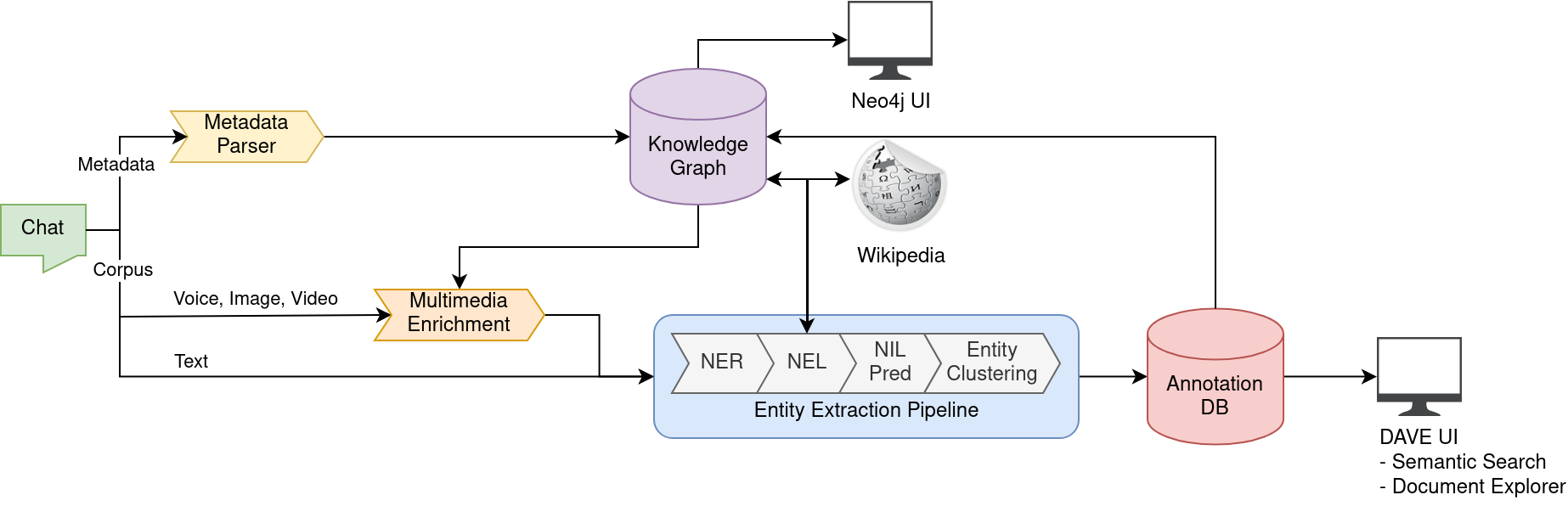}
\caption{Overview of the proposed solution.} \label{fig:overview}
\end{figure}



A sketch of our solution is depicted in Figure~\ref{fig:overview}. It combines different technologies: KGs to organize data extracted from IMA dumps, deep learning-based speech-to-text technology to transcribe audio messages and enrich chat-specific data, NLP to extract entities and annotate chat data, and semantic search to access the data. 
Chat dumps include the content of the messages and their metadata; the latter include, e.g., participants, message timestamps, type of message, and so on. The dumps are processed to populate a property graph, i.e., the chat KG, stored in Neo4j\footnote{https://neo4j.com/}. The graph represents relational data (chats, participants, exchanged messages) as well as the content of the messages. Multimedia enrichment generates textual content for audio messages and other attachments; this is intended to expose the data to search functionalities and enable entity extraction and subsequent annotation of the chats.  
In the current version, we focus on transcribing audio messages with the Whisper APIs~\cite{radford23a-whisper}; however, in the figure, we also include hints on natural extensions of the multimedia enrichment steps (image-to-text, and video-to-text enrichment). Entities appearing in the chat metadata (senders, participants) are implicitly tracked when the graph is built or found with deterministic rules from input data. For the content of the messages, we use an entity extraction pipeline based on a Named Entity rEcognition and Linking (NEEL) pipeline adapted from previous work~\cite{pozzi2023,pozzi23aixia}.
It combines Named Entity Recognition (NER), Named Entity Linking (NEL), NIL prediction 
and entity clustering (for more details, see Section~\ref{sec:approach}).  
Based on NEEL, we annotate the chat dumps serialized into text files, which form an annotated corpus, and we update the KG with information about the extracted entities.   

In this paper, we consider the two main User Interfaces (UI) developed to access the data and get insights: 
the first one is a native Neo4j UI that 
supports graph queries and visual exploration of the graph 
(the opportunity to exploit this feature was a major driver in choosing Neo4j for storage, considering the effort available in the project). The second one is based on DAVE (Document Annotation Validation and Exploration), a web application we are developing to visualize, search, and edit annotated documents based on faceted search, where users can filter results using entity-based annotations. 



This paper is structured as follows: in Section~\ref{sec:context} we contextualize our work and discuss the relevance of IMA for criminal investigations; in Section~\ref{sec:relatedworks}, we discuss related work. Section~\ref{sec:approach} describes in detail the components of our approach and Section~\ref{sec:results_and_challenges} discusses experimental results and challenges. Conclusions are presented in Section~\ref{sec:conclusion}.
Finally, we report our ethical considerations in Section~\ref{sec:ethics}.

\section{Project Context and Challenges in the Analysis of IMA} \label{sec:context}

The broader scope of the project this work is part of is to apply semantic integration methodologies to manage the large amount of data collected in the investigation~\cite{batini2021semantic}. This data comes from a heterogeneity of sources such as documents, e.g., reports by consultants, (semi-)structured files, e.g., phone call records, bank transfers, and digitalized papers, e.g., seized in the investigation. As a result of the previous interactions, we had the first positive feedback on the overall semantic integration concept and defined a reference architecture~\cite{batini2021semantic}. In a later stage,  
we focused on more specific questions: which aspects of their activities can be impacted more positively by the use of AI and semantic integration? 

The first aspect we identified is the need to \textbf{integrate the pieces of evidence} collected directly and indirectly. During the investigation, before the decision of formulating or dropping charges for the trial follow-up, and during the preparation for the trial, prosecutors need to search in the documentation, reason on top of it, and use it for structuring an argument. Even when different specialized products are used, integration across pieces of evidence
are processes where available software products offer no or very limited help.   

A second aspect we identified is the need to \textbf{quickly access content whose analysis represents a critical barrier} because of the associated costs (time and human effort). Such costs, in the best case,  impact dramatically on the length of the investigation, and, in the worst case, prevent them from extending the analyses to the scale they deem useful (budget allocated to investigations may change also depending on the severity of the charges).
%
An effort-demanding activity that is considered particularly important in investigations is the analysis of data collected from IMA on mobile phones seized during an investigation, which we consider in this paper.  




IMA data found in seized mobile phones 
provide valuable insights into the owner's communication network, and enclose messages and attachments (images and audio and video files) that often can be used as evidence or hints for supporting, rejecting, or triggering investigative hypotheses. 
In some investigations, law enforcers 
use a software product to inspect its whole content, which helps users explore data from a user interface and perform textual searches. However, investigators found that the knowledge exploration process is still quite limited because searching is based on syntactic patterns only, and no integration support is offered. They also know products on the global market with richer features, 
but reported that the costs of licenses are incompatible with most of the investigations.    
We started to unlock the content of audio files, because, according to investigators, they are often too much resource-demanding to process. An upper bound estimate of the time required to transcribe 1h of speech is 8h of work.

The work discussed in this paper was conducted in two stages using data from two investigations. In the \textit{first stage}, we extracted a graph from the initial investigation's data, developed graph-based querying and visualization tools, and produced a report addressing specific questions posed by the investigators as the final result of our consultancy contract. In the second stage, we tested speech-to-text processing and entity extraction. The expected outcomes of our work consist, indeed, of a report and a PoC to demonstrate the proposed solutions.


While factual analysis may target facts that investigators believe may have happened, it may also target the rejection of claims of the investigated subject. For example, in one case the Person Under Investigation (PUI) claims that he had scheduled a meeting with another subject, but investigators did not believe that the claim was grounded; while the subject could not prove that the meeting had happened, we were asked to check at the best of our capacity traces of such an encounter, to support or reject the argument that no evidence of such meeting is found in the mobile phone. These questions highlight two important ethical aspects of having AI tools in the loop:
\begin{itemize}
    \item It is important to make sure that the evidence extrapolation process can be traced, controlled, and explained, in such a way that the evidence can be used by investigators for decision-making and, later on, in the trial; as a result, we believe that empowering search is the safest approach to propose today.  
    \item It is important to design solutions that consider the limitations of the adopted models and tools by design, and the intrinsic uncertainty associated with data analysis; as a consequence, our PoC tools are designed to make it possible to modify annotations predicted by algorithms, support syntactic keyword search as a base functionality and help users understand the quality of the results.   
\end{itemize}

As such, we remark again that the final objective of the target solution is by no means to substitute the human decision-maker but only to empower her/him with better tools to explore the evidence collected within the investigation so as to build more robust arguments and better defend in the trial.

In relation to the project objectives summarized in Section~\ref{sec:introduction} (PoC development and feature specification), we remark that our goal is to test,     
collect feedback from the investigators about the usefulness and fit for use of the proposed solution, and better focus the identified features to be developed, rather than engineering a final product. 

\section{Related Work} \label{sec:relatedworks}

As discussed in Section~\ref{sec:introduction}, this work is an evolution of the approach described in~\cite{batini2021semantic} and is inspired by previous work to combat human trafficking based on semantic data integration~\cite{szekely2015building}. With respect to~\cite{batini2021semantic}, we specifically target IMA data, consider multimedia data enrichment, use a substantially improved NEEL pipeline, and include a semantic search interface. From~\cite{szekely2015building}, we borrow the overall entity-centric approach~\cite{kejriwal2018}, as well as the idea to improve search as a main objective~\cite{kejriwal2017investigative}. However, we work on IMA data with a native network structure rather than on data crawled from the web, and we apply a full-fledged NEEL pipeline to extract and interlink entities. Direct exploitation of graph queries and multimedia data enrichment are two additional novelties. 

Other approaches exploit knowledge graphs to detect criminal activities in the financial domain, employing various AI techniques and logic-based KGs and reasoning~\cite{alhajeri2023using,bellomarini2020rule}. However, they operate on non-textual or non-speech data. Another semantic approach studies forensic ontologies and reasoning, but without discussing knowledge extraction techniques~\cite{spyropoulos2023interoperability}.

      


Several NLP approaches have been developed for the broader legal domain, but most of them do not adopt KGs and a semantic data integration paradigm.
NER and relation extraction have been proposed in the broader legal domain~\cite{zhong2020does}, and also to support forensic analysis~\cite{IE-forensic-2015}. The latter work, in particular, does not focus on IMA or apply a full NEEL pipeline, but exploits visualization as an intermediate step to check extracted entities before extracting relations. This usage of a human-in-the-loop approach inspires the design of our DAVE application and its future evolution. 
In conclusion, there are several novelties with respect to the application of NLP to criminal investigations, from the application to IMA data, to the usage of a NEEL pipeline.  


Concerning multimedia enrichment and analysis, a work focused on images and speech for crime prevention, applying speech-to-text models and NLP techniques~\cite{P_rez_2021}; however, it emphasizes behavioral pattern detection and sentiment analysis, rather than entity extraction.



In relation to the NEEL pipeline used in this paper for extracting entities, 
we refer to papers where these algorithms have been discussed more in detail~\cite{pozzi2023,pozzi23aixia}. In these references, we also provide a more precise comparison between our algorithms and other algorithms used in the legal domain, and we discuss some limitations of our pipeline. 
Entity clustering has been found as a less explored topic, which is currently under the attention of the NLP and semantic web communities~\cite{logan2021benchmarking,Heist2023nastylinker,kassner-etal-2022-edin}. In our work, we have been inspired by~\cite{kassner-etal-2022-edin}, but we have been trying to improve it with additional features to compute mention-mention similarities and with community detection algorithms. A quantitative comparison between our clustering approach and previous work is out of the scope of this paper. 

Broadening the context, papers, and books reviewing techniques for building Knowledge Graphs (KGs) and domain-specific KGs~\cite{weikum2021machine,ZhaoMultisourceFusion,KejriwalBook} are relevant to the domain-specific work discussed in this paper. They stress the importance of combining different techniques for information extraction, including handcrafted rules to capture domain specificity~\cite{weikum2021machine}. 

\section{Approach} \label{sec:approach}

To provide more details about our approach, we describe the extraction of the knowledge graph (Section~\ref{sec:metadata_kg}), the methodology we adopt for multimedia (Section~\ref{sec:multimedia}), the entity extraction pipeline (Section~\ref{sec:entity_extraction}), and the exploratory interfaces (Section~\ref{sec:exploration}).



        
    \subsection{Organizing the conversations with the knowledge graph} \label{sec:metadata_kg}
Chat dumps are available in a copy of the mobile phone extracted with certified forensic applications. 
Information about each chat includes the list of participants with their phone numbers, the start time, and the time of the last activity. Each message is described by its timestamp, the name and number of the sender, and the attachments it contains.
We consider these metadata to be applicable to most instant-messaging applications. 

However, different forensic applications are likely to export data in different formats. In our case, we received structured textual data\footnote{We used parsing expression grammars (PEGs) from https://pypi.org/project/parsimonious/.} for the first investigation and MS Excel (.xlsx) data for the second one.



\begin{figure}[h!]
\centering
\includegraphics[width=0.7\textwidth]{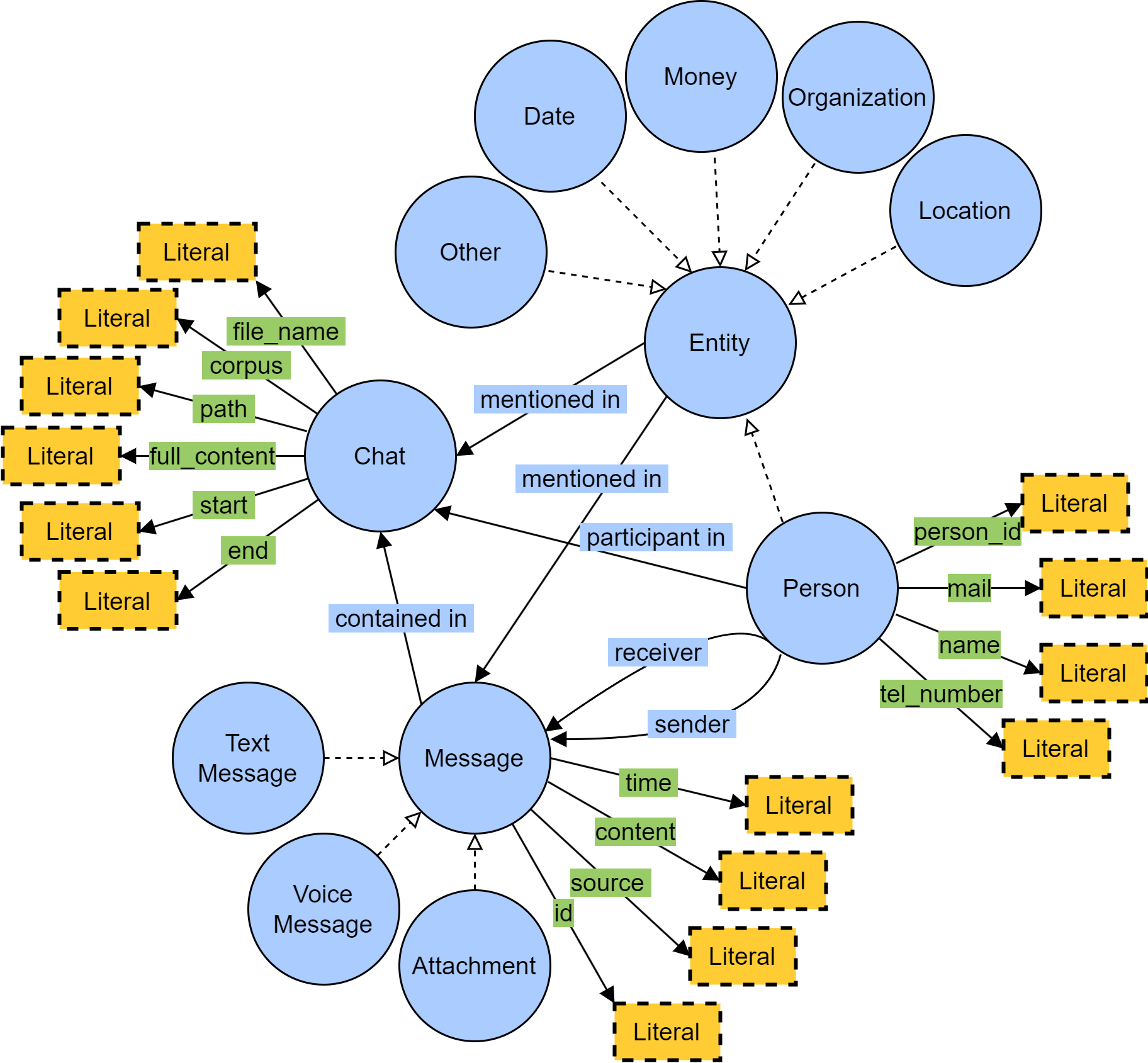}
\caption{Graph schema.} \label{fig:model}
\end{figure}

Once extracted, metadata and messages contents are used to populate the Neo4j chat knowledge graph. A conceptual schema of the knowledge graph is shown in Figure~\ref{fig:model}; dashed arrows represent is-a relations, continuous arrows represent relations and properties. We also represent \textit{sameAs} relationships, which are extracted from the chat metadata (different numbers associated with the same contact).      
The main reason to choose a property graph and Neo4j over semantic web standards is the opportunity to exploit its visual interface for reporting purposes.     


    
    \subsection{Multimedia data enrichment} \label{sec:multimedia}


%

While we plan to increase the support for multimedia content, our current pipeline is designed to process audio files only, since, according to the prosecutors we collaborated with, this is the most time-consuming type of multimedia to deal with. 

To transcribe audio files we utilized Whisper~\cite{radford23a-whisper}, an automatic speech recognition (ASR) system developed by OpenAI and implemented as an encoder-decoder Transformer. 

The decision to utilize Whisper as our go-to algorithm for speech-to-text came after an evaluation of the performances of several Automatic Speech Recognition (ASP) systems. For this evaluation, we randomly selected 500 Italian audio files and their respective validated transcription from three sources: Mailabs\footnote{https://www.caito.de/2019/01/03/the-m-ailabs-speech-dataset/}, CommonVoice\cite{ardila2019common} and VoxForge\cite{Voxforge.org}. Since these audio have, in general, exceptional quality, we decided to add artificial noise, in order to better reflect the quality expected in WhatsApp's voice messages, which usually contain various types of environmental noise. To do so, we utilized different audio distortion techniques, such as gaussian noise, background noise, speed up, pitch shift, delay, and distortion. These transformations were applied randomly to the audio files. We then evaluated the performance of the systems with two well-known measures: word error rate (WER) and BERT score (F1 score). The best-performing model was found to be Whisper (Large), which obtained a BERT score equal to 0.908 and a WER score equal to 0.282.


\subsection{Algorithms for entity extraction} \label{sec:entity_extraction}

\begin{figure}
\includegraphics[width=\textwidth]{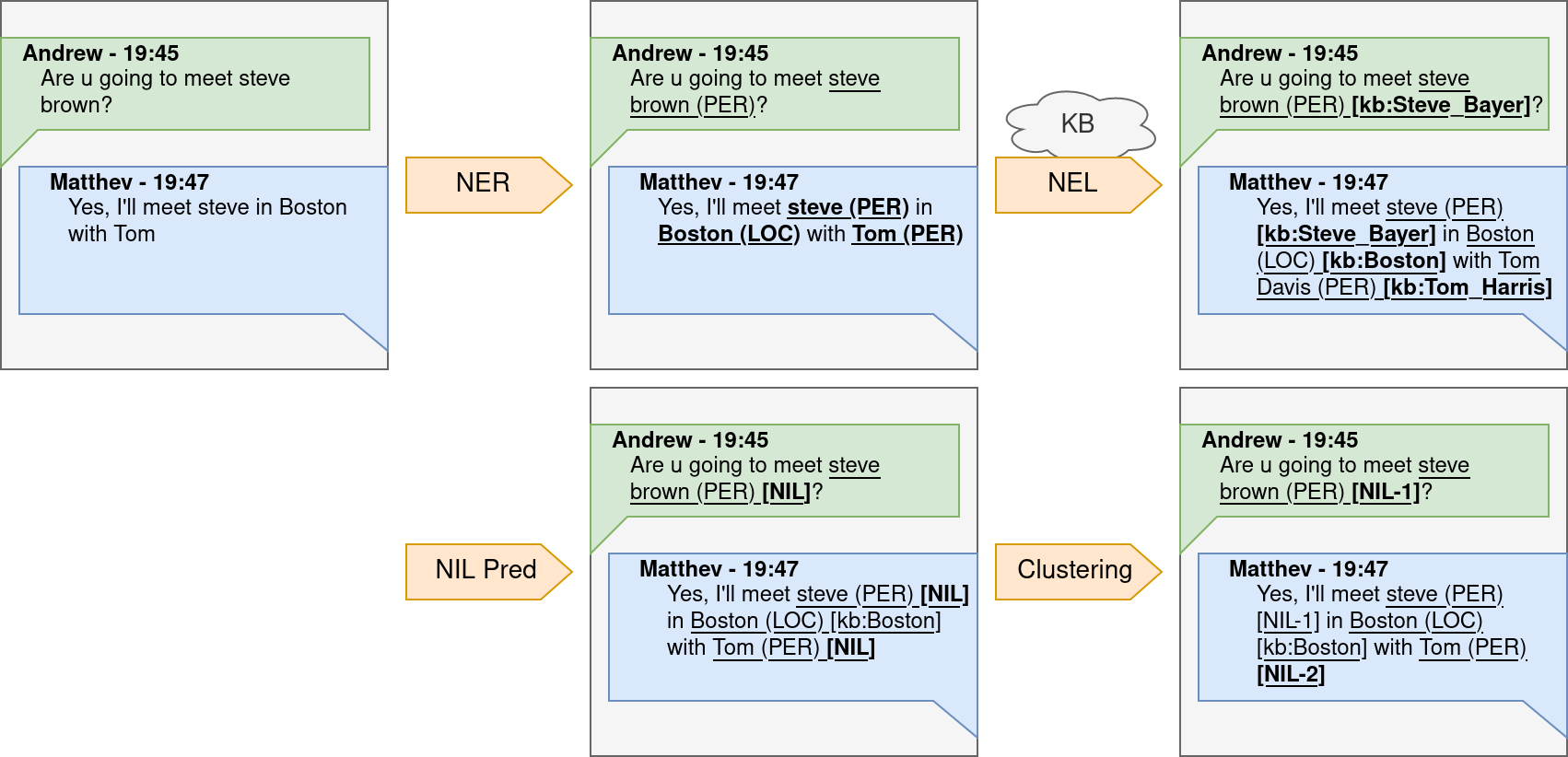}
\caption{The entity extraction pipeline on chat messages. NER identifies mentions of entities,
NEL identifies candidate entities from the knowledge graph (KG). NIL prediction detects that \underline{steve brown}, \underline{steve}, and \underline{Tom} are linked to wrong entities and assumes they are not in the KG (NIL), while \underline{Boston} is linked to the KG. Entity clustering assigns \underline{steve brown} and \underline{steve} to the same cluster (\underline{NIL-1}), as they refer to the same entity, and \underline{Tom} to another cluster (\underline{NIL-2}).} \label{fig:neel}
\end{figure}

Most of the pipeline components for entity extraction directly derives from our previous  work~\cite{pozzi23aixia}, 
while the entity clustering used is specific to this study. 
Figure~\ref{fig:neel} depicts an example application of applying the NEEL pipeline on chat messages.
In the remainder of this section, we describe each component of the NEEL pipeline.

The NER component is based on the library SpaCy-transformers\footnote{https://spacy.io/universe/project/spacy-transformers} with the SpaCy transition-based parser to leverage contextualized token representations obtained from a transformer~\cite{Vaswani2017}. The transformer we use is based on ITALIAN-LEGAL-BERT~\cite{licari_italian-legal-bert_2022},
additionally fine-tuned with masked language modeling (MLM) 
for the Italian legal domain and finally for NER 
in~\cite{pozzi23aixia}).



For NEL, we use the bi-encoder architecture of BLINK~\cite{wu-etal-2020-scalable}, trained for the Italian language using Italian Wikipedia\footnote{https://it.wikipedia.org} hyperlinks. 
As the knowledge base for NEL we use the entities from Italian Wikipedia (without redirects and disambiguation pages) combined with the participants of the chats previously organized in the chat knowledge graph.
Specifically we exploit BLINK zero-shot capability to encode an entity given its title and textual description, thus for each participant we obtain a representation giving the following input to the bi-encoder.
\begin{verbatim}  
    [CLS] {name} [ENT] phone number: {phone1(,phone2,...)} [SEP]
\end{verbatim}


In this work, since the vast majority of the people of interest for the investigation refers to entities not in Wikipedia we link mentions of persons only to the chat knowledge graph. For organizations and locations, instead, we consider Wikipedia entities.

In the NIL prediction component, we employ a \textit{logistic regression classifier}. 
It takes two inputs: 1) the NEL system's top-ranked entity score and 2) the score difference between the top-ranked entity and the second-best one. 
The output $p \in [0,1]$ denotes correctness for linking the mention (1) or NIL (0) if the correct entity isn't the top-ranked.

The entity clustering approach is based on a supervised xgboost classifier trained on pairs of mentions belonging to the same cluster, determined through NEL data from the gold standard. This classifier takes as input a set of similarities, including surface text similarities (like Jaro Winkler and Jaccard), as well as cosine similarity of BLINK embeddings obtained with NEL, producing a probability score as an indicative synthetic similarity between the mentions. The pairwise similarity scores are then used to create weighted graphs for each document, where mentions are nodes and edge weights reflect mention similarities. The final mention clusters are obtained by applying a community detection algorithm (Louvain method~\cite{blondel2008fast}) on these graphs.

The application of NEEL has two outcomes: the chat file is annotated
and the knowledge graph is updated with the discovered entities using the ``mentioned in'' relationship (see Figure~\ref{fig:model}).


\subsection{Exploration and editing of annotated data} \label{sec:exploration}


To query, explore and get insights into the processed data, we provide two main interfaces: querying and exploring the knowledge graph stored in Neo4j, searching and exploring the chat files with faceted search.

\begin{figure}
\includegraphics[width=\textwidth,trim=0 140 0 100, clip]{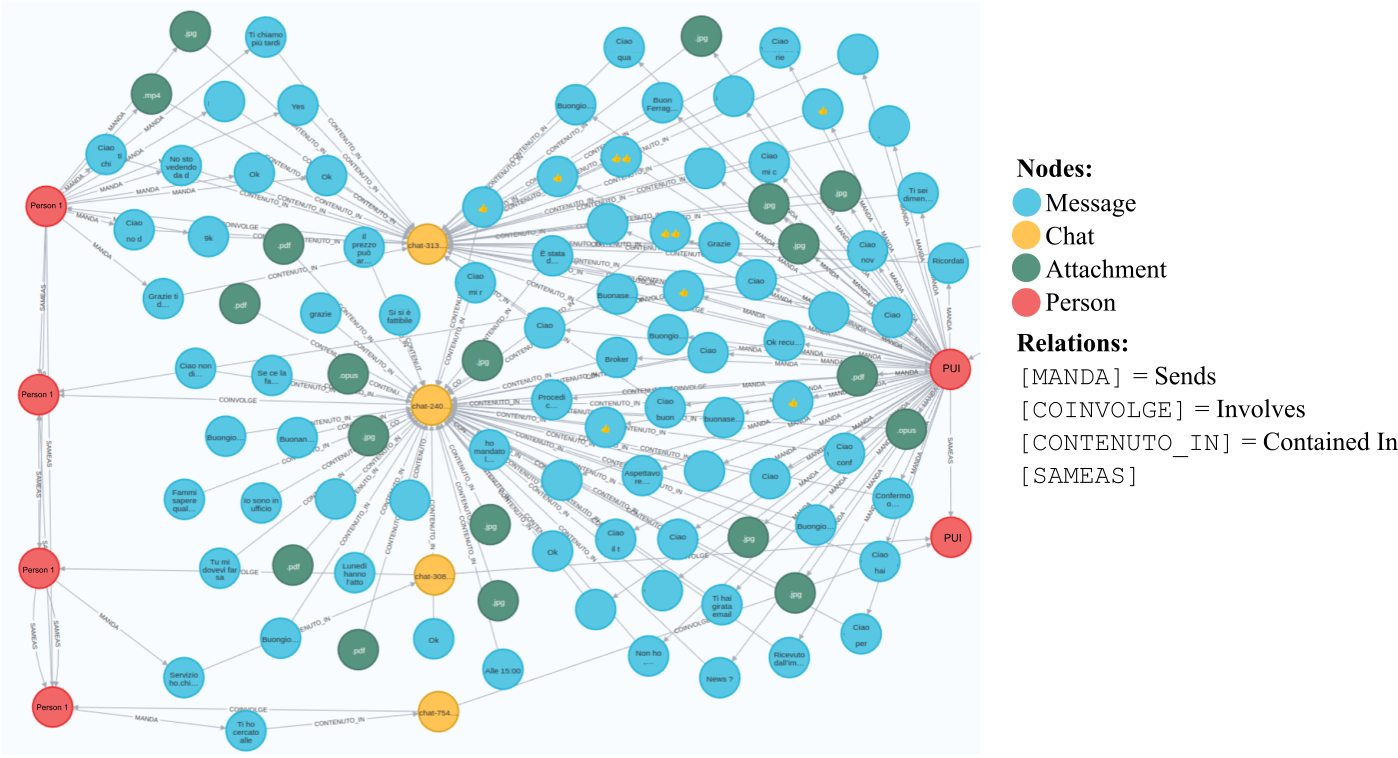}
\caption{Analysis of the correspondences between Person Under Investigation and a second individual. We provide the translation for relationships in Italian.
} \label{fig:KGexploration}
\end{figure}

\textbf{Knowledge graph exploration.}
Users can navigate the knowledge graph created by metadata using the query and visual exploration interface of Neo4j. This approach provides a comprehensive overview of the relationships and connections between various elements in the investigative data, but it also supports specific queries on the data. An example of this interface can be seen in Figure~\ref{fig:KGexploration}. The Figure shows all messages and chats between the Person Under Investigation (PUI) and another subject; sameAs connections reveal that the PUI and the other subjects have different numbers used in different chats.    

\begin{figure}
    \begin{subfigure}[b]{0.49\textwidth}
        \centering
        \includegraphics[width=\textwidth, trim=0 0 270 0, clip]{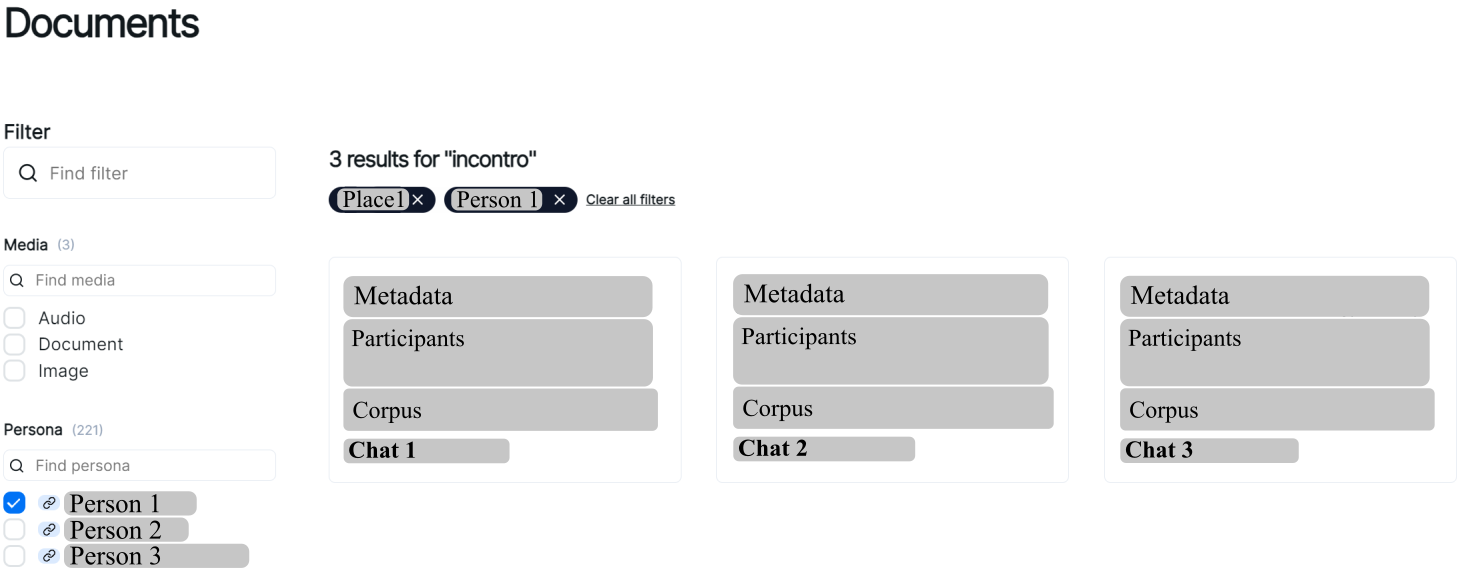}
        \caption{Using the Faceted Search on DAVE to retrieve relevant information. Sensitive information has been obscured.} 
        \label{fig:Faceted Search}
    \end{subfigure}
    \hfill
    \begin{subfigure}[b]{0.49\textwidth}
        \centering
        \includegraphics[width=\textwidth, trim=0 0 100 0, clip]{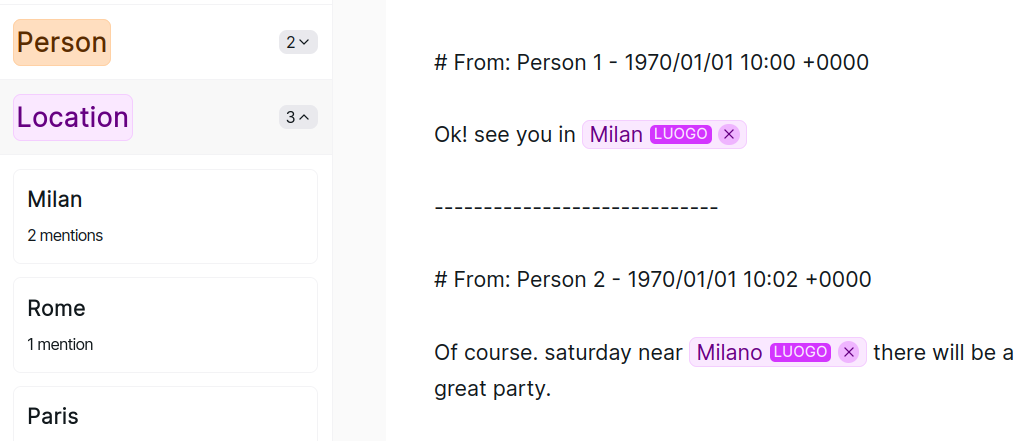}
        \caption{Using the Document Explorer to navigate the entity mentions in the chat.} 
        \label{fig:document explorer}
    \end{subfigure}
\end{figure}

\textbf{Faceted Search and Document Exploration.} Faceted search allows users to retrieve chats through a traditional keyword search. Users can filter results by selecting specific metadata (e.g., participants) or semantic annotations (e.g., other people, organizations, or places mentioned in the chat). Our current implementation is based on Elasticsearch\footnote{https://www.elastic.co/elasticsearch}. A peculiar feature is that semantic search is aware of links and clusters identified in the annotation process. 
A sanitized example of the Faceted Search interface is shown in Figure~\ref{fig:Faceted Search}; three chats containing the keyword ``incontro" (Italian for ``meeting") are found; these chats also contain specific persons and specific places shown in the panel on the left; by selecting a specific entity or metadata, the results are filtered. A user can click on a chat she/he wants to open using the \textbf{Document Explorer} page; from this interface, as shown in Figure~\ref{fig:document explorer}, the user sees the clusters of entities in the left panel and all their mentions; by clicking on a mention, she/he can move to the exact position of the mention. 



\subsubsection{Role of Different Interfaces and Annotation Editing}

While the search patterns supported by the two interfaces overlap, each interface has a specific role. 
Table~\ref{tab:interfaces role} summarizes the information each interface is able to provide the user, in which form, and the impact of the graph representation and of the NLP entity extraction to the results given to the user.


Additionally, DAVE was designed for a human-in-the-loop approach, considering the sensitive nature and ethical implications of the legal domain. Indeed, users can edit the annotations, correcting them, ensuring the quality of the extracted knowledge.



\begin{table}[h]
\footnotesize
\begin{tabular}{|p{1.3cm}|p{1.7cm}|p{1.4cm}|p{3.8cm}|p{3.5cm}|} 
\hline
Interface &
  Query &
  Result &
  Impact of graph data &
  Impact of NEEL \\
  \hline
  \hline
Graph\newline Interface &
  Cypher
  &
  Tabular, Graph &
  Match all nodes attributes. Graph visualization &
  Queries match ``mentioned entity'' nodes \\
  \hline
DAVE &
  Faceted Search &
  Chats &
  Annotations linked to entities in the chat knowledge graph &
  The entities occurring in the matching chats:\newline - are shown to the users\newline- can be used to filter \\
  \hline
DAVE &
  Document Exploration &
  A specific position in a chat &
  Annotations linked to entities in the chat knowledge graph &
  The entities occurring in the chats:\newline - are shown to the users\newline - can be used to filter\\
  \hline
\end{tabular}
\caption{Role of different interfaces.}
\label{tab:interfaces role}
\end{table}

        
\section{Experimental Results, Challenges, and Limitations} \label{sec:results_and_challenges}
The results discussed in this section reflect the two-stage development of the project (see Section~\ref{sec:context}) and the maturity gap between the extraction of the enriched graph (first project stage) and the application of the NEEL pipeline (second project stage). Since the two investigations cover different domains, we consider data from both of them, focusing on the second one for specific analyses.  

Our evaluation has three main objectives: 1) to provide evidence that the proposed combination of graph-based modeling, multimedia enrichment, and NLP-based entity extraction can improve the analysis of IMA data in criminal investigations by exploiting graph-based queries and the semantic search paradigm; 2) to provide insights into the quality of the proposed solution in its current development stage; 3) to discuss current limitations and challenges.       


\begin{table}[htbp]
\small
  \centering
    \begin{tabular}{|l|r|r|r|r|r|}
    \hline
    \textbf{Investigation} & \textbf{Chats} & \textbf{Proc. chats} & \textbf{Messages} & \textbf{Attach. (img-audio-docs)} & \textbf{Persons} \\
    \hline
    1) Fraud & 1133  & 801   & 45252 & 3324 (575-304-1590) & 1365 \\
    \hline
    2) Corruption & 1442 & 1442  & 364690 & 63224 (51532-6273-4066) & 2351 \\
    \hline
    \end{tabular}  \caption{Investigation Statistics}
  \label{tab:chatstats}
\end{table}

\textbf{Graph construction.}
Statistics that summarize the result of chat processing to construct the graph are reported in Table~\ref{tab:chatstats}. For each investigation, we report its topic and the number of input chats, successfully processed chats loaded in the graph, messages, attachments (with more details about images, audio and documents), and number of different persons involved in the chats. Chats may involve two persons or groups (e.g., 86 groups in the second one).  

\textbf{Impact of entity-based modeling on search.} 
In the first investigation, we used the tool to write a report. Therefore, we translated each investigative question into Neo4j queries and reported the findings, providing: the queries, arguments and explanations for their choice, graph-based visualizations provided by the UI (e.g. Figure~\ref{fig:KGexploration}), and copies of the relevant messages. We can classify the query patterns into three main groups based on their purpose: relational analysis, text search, and entity search (see examples in the supplementary material).   

We found out that keyword and entity searches were very frequent. After validating the perceived utility of combining graph-based constraints and textual search\footnote{This combination makes it possible to visualize the nodes (messages) that contain the indicated word and to navigate the graph, displaying the nodes connected to one of interest.} and including graph-based visualizations in reports, in the second investigation we dug deeper into the impact of our entity-centric approach to empower search at a larger scale (messages and transcripts increased by an order of magnitude in the second investigation). Law enforcers handed over a list of 30 keywords of interest (22 generic terms and 8 entity names) to search for, for evaluation purposes.
The total number of results returned by searching each keyword in the list amounts to 818 chats, 1780 messages, and 230 audio transcripts (results for different keywords may overlap). 
The generic terms\footnote{We use here their English translation} returning more results (chats / messages) are ``agreement'' (126 / 274), ``runoff'' (113 / 133) and ``tender'' (92 / 296).  These numbers suggest that the results space for keyword search is high, motivating the need for more advanced mechanisms to filter out results and explore the data specifying constraints on specific entities. For each query, we count how many distinct entities are present in the retrieved chats. We distinguish between the entities extracted by IMA metadata (without considering the message corpus) and the entities recognized by the NEEL pipeline from messages' corpora, more prone to errors. The median number of distinct entities retrieved from chats metadata 181, and 568 for the entities recognized by the NEEL pipeline. These entities can be used to filter out the often large number of results (e.g., consider only messages sent by X or in chats with Y) or provide insights into entities relevant to the query context. The amount of data and associations between entities and results support the introduction of the faceted search and document explorer interfaces, which display the entities found in the results.   

The terms referring to entities (persons and organizations) return fewer results (max 56 chats, min 0). Two terms are found in audio transcripts, providing evidence for the impact of the multimedia enrichment. We also match these terms to names of distinct entities extracted from the messages' content with the NEEL pipeline or included in the graph (w/o considering the NEEL pipeline). While three entities are found in both sources, four are found in only one source, and one in none. We believe that this observation suggests that combining keyword search with filters based on entities identified with two criteria (rule-based graph construction and NEEL-based extraction) is beneficial (more details are provided in the supplementary material).

\textbf{NEEL contribution to knowledge graph enrichment.} The NEEL pipeline successfully processed 1296 out of 1442 input chats, with errors caused by technical issues that can be fixed by splitting the very long chats. Statistics about the extracted entities are reported in Table~\ref{tab:stats}: the table indicates the impact of NEEL on enriching the KG with the ``mentioned in'' relationship, 
and the informativeness of audio transcriptions, which contain on average four times more entities than text messages. 
The statistics also show that 2361 mentions of persons have been linked to entities in the chat knowledge graph, which validates the introduction of this in-domain linking mechanism.

\begin{table}[htbp]
  \begin{minipage}[t]{0.45\linewidth}
    \centering
    \begin{tabular}{|c|c|c|c|c|c|}
      \hline
      & \multicolumn{2}{c|}{\textbf{\#Mentions}} & \multicolumn{2}{c|}{\textbf{\#Links}} & \textbf{\#Entities} \\
      \hline
      & \textbf{Text}  & \textbf{Audio} & \textbf{to KB} & \textbf{NIL} & \\
      \hline
      \hline
      Per & 7765 & 520 & 2361* & 5404 & 5701 \\
      Org & 1753 & 113 & 614 & 1139 & 1339 \\
      Loc & 2578 & 185 & 1118 & 1460 & 1892 \\
      Date & 916 & 53 & - & - & - \\
      Money & 124 & 23 & - & - & - \\
      Misc & 32 & 1 & - & - & - \\
      \hline
    \end{tabular}
    \caption{Statistics of the NEEL extraction. * Links for Person refer to the chat knowledge graph, not to Wikipedia.}
    \label{tab:stats}
  \end{minipage}%
  \hfill
  \begin{minipage}[t]{0.45\linewidth}
    \centering
    \begin{tabular}{|c|c|c|c|}
      \hline
      & \textbf{Prec} & \textbf{Rec} & \textbf{F1} \\
      \hline
      \hline
      Strong & 44.0 & 21.4 & 28.8 \\
      Partial & 71.4 & 35.1 & 47.0 \\
      \hline
    \end{tabular}
    \caption{NER Evaluation. ``Strong'' counts perfect matches of both span and type, while ``Partial'' also counts as correct when there is an overlap between predictions and gold standard annotations.}
    \label{tab:eval}
  \end{minipage}
\end{table}

\textbf{Preliminary insights on the quality of NEEL annotations.}
While an end-to-end evaluation of the NEEL output is ongoing, we discuss a first evaluation of the NER component. 
We annotated a gold standard for evaluating NER derived from six chats of which three are group chats. We selected the chats to annotate among the ones that are 
long enough and contain audio transcripts. 
We computed a first set of automatic annotations using the NER component and thoroughly revised them with Doccano\footnote{https://github.com/doccano/doccano}. At the end of the process, we annotated 668 mentions of Person, 268 of Organization, 207 of Location, 157 Dates, and 11 Money.

Results shown in Table~\ref{tab:eval} are not satisfactory, confirming that NEEL on IMA data is challenging because of the specific data distributions and suggesting that in-distribution fine-tuning is necessary. 
Indeed IMA data contain peculiarities such as person name abbreviations or inconsistent capitalization of names and locations. In relation to fitness for use, we make two remarks. 1) Metadata are precisely identified enabling accurate filtering in DAVE, for instance, by message sender. 2) DAVE supports text search, which makes it possible to complement entity-based retrieval to improve result recall.

\textbf{Investigators' Feedback}. The feedback on our work was collected based on the report and demonstrations of our PoC (live demo) across the collaboration stages. 
Investigators found the graph-based visualizations helpful and understandable. They are familiar with these structures because of the relevance of relational knowledge in their investigations. The combination of text search with relational queries was very appreciated. In relation to multimedia data enrichment, they believe that speech-to-text (and similar enrichment technologies) could be really transformative in terms of speed and effort balance.
They have been quite surprised by the quality of the transcriptions and not concerned by occasional mistakes, as they always verify the evidence against the original audio files. In fact, they suggested adding audio snippets to the UI for direct listening.
The NLP part was also appreciated in terms of potential and usefulness. The opportunity to quickly browse through the chat exploiting DAVE interface was deemed very promising because their previous tools only supported syntactic match patterns.


\textbf{Limitations and challenges.}
Our current solution has several limitations. To overcome the weaknesses of pre-trained algorithms in the NEEL pipeline, we need to increase the size of the gold standard and fine-tune the algorithms on IMA data. 
%
After our demonstrations, investigators were eager to use our solution, but querying Neo4j through its interface requires expertise, and expressed their interest in conversational search interfaces to answer questions. We therefore developed two additional prototype interfaces 1) "a set of intuitive and composable Python functions that handle a wide range of common query patterns,
and 2) a conversational search interface based on the popular paradigm of retrieval augmented generation. We did not describe these features in this paper as their development stage is too early. However, the prosecutor and the law enforcers actually accessed the simplified query interface and were able to submit queries.    

\section{Conclusion} \label{sec:conclusion}


In the paper, we presented an application of knowledge graphs to model data extracted from IMA in the context of criminal investigations. In addition to offering graph-based query access to the data, we used speech-to-text to enrich the original data with transcriptions from audio messages. Finally, we applied NLP to extract entities and support semantic search in a prototype application. The work has been developed in collaboration with investigators and experimented in two real-world investigations. The experiments discussed in the paper and the feedback collected from the investigators provide evidence for the \textit{usefulness} of the proposed solution, the \textit{soundness of the main technical choices}, and the applicability to \textit{realistic data volumes}. They also highlights limitations, which we believe can be addressed with a reasonable amount of effort. 

In future work, we plan to improve the NEEL pipeline through in-domain fine-tuning, continue to develop and evaluate the interfaces (especially the new ones), and extend data enrichment to images and videos.     


\section{Ethical Considerations} \label{sec:ethics}
Our work is formalized through an official investigation consultant contract. Data is stored on local servers, with access restricted to secure VPN connections and user-password authentication. To safeguard private information, we have avoided using external APIs. Finally, the solution proposed in this work features are designed to allow accurate verification of information against the original data.

\bibliographystyle{splncs04}
\bibliography{bibliography}

\end{document}